\documentclass[11pt,a4paper]{article}
\usepackage[hyperref]{acl2018}
\usepackage{times}
\usepackage{latexsym}
\usepackage{graphicx}  
\usepackage{amsmath}
\usepackage{color}
\usepackage{subfig}
\usepackage{multirow}
\usepackage{enumitem}

\usepackage{url}
\usepackage{arabtex}
\usepackage{utf8}
\setcode{utf8}

\aclfinalcopy 

\title{Diacritization of Maghrebi Arabic Sub-Dialects}

\author{Ahmed Abdelali$^{\ast}$, Mohammed Attia$^{\star}$, Younes Samih$^{\dagger}$, Kareem Darwish$^{\ast}$ and Hamdy Mubarak$^{\ast}$\\
  $^{\ast}$Qatar Computing Research Institute, $^{\star}$Google Inc., $^{\dagger}$University of D\"usseldorf\\
$^{\ast}$aabdelali@hbku.edu.qa, 
          $^{\star}$attia@google.com,
            $^{\dagger}$samih@phil.hhu.de \\ $^{\ast}$kdarwish@@hbku.edu.qa, $^{\ast}$hmubarak@hbku.edu.qa
           \\}

\date{}

\begin{document}
\maketitle
\begin{abstract}
Diacritization process attempt to restore the short vowels in Arabic written text; which typically are omitted.
This process is essential for applications such as Text-to-Speech (TTS). While diacritization of Modern Standard Arabic (MSA) still holds the lion share, research on dialectal Arabic (DA) diacritization is very limited. 
In this paper, we present our contribution and results on the automatic diacritization of two sub-dialects of Maghrebi Arabic, namely Tunisian and Moroccan, using a character-level deep neural network architecture that stacks two bi-LSTM layers over a CRF output layer. The model achieves word error rate of 2.7\% and 3.6\% for Moroccan and Tunisian respectively and is capable of implicitly identifying the sub-dialect of the input. 
\end{list}
\end{abstract}

\section{Introduction}

Arabic is typically written without diacritics (short vowels)\footnote{List of Arabic diacritics: fatha (a), damma (u), kasra (i), sukun (o), shadda ($\sim$).}, which require restoration during reading to pronounce words correctly given their context. For MSA, diacritics serve dual function, namely: word-internal diacritics 
dictate pronunciation and lexical choice; and end of word diacritics (aka case endings) indicate syntactic role.  Conversely, dialects overwhelming use \textit{sukun}, which typically indicates the absence of a vowel, as case endings, eliminating the need for syntactic disambiguation. Thus, DA diacritic recovery mostly involves restoring word-internal diacritics. Diacritic restoration is crucial for applications such as text-to-speech (TTS) to enable the proper pronunciation of words.  Though sub-dialects could be orthographically identical, regional phonological variations necessitate specific tuning for sub-dialects. 

In this paper we present new state-of-the-art Arabic diacritization of two sub-dialects of Maghrebi, namely Moroccan (MOR) and Tunisian (TUN).  We employ a character-level Deep Neural Network (DNN) architecture that stacks two bi-LSTM layers over a Conditional Random Fields (CRF) output layer. The model achieves word error rate (WER) of 2.7\% and 3.6\% for MOR and TUN respectively.  Further, the model is capable of implicitly identifying the sub-dialect of the input enabling joint learning and eliminating the need for specifying the sub-dialect of the input.  We compare our approach to an earlier work based on CRF sequence labeling~\cite{DARWISH18.20}. Our contributions are: 
\begin{itemize}[leftmargin=*]
\item Our novel work on Maghrebi diacritization shows some traits of Maghrebi (e.g. effective out-of-context diacritization) and provides strong results.
\item Improve earlier results of using CRF.
\item We explore cross dialect and joint training between MOR and TUN. Our DNN approach can effectively train and test on multi-dialectal data without explicit dialect identification.
\end{itemize}

\section{Background}
Most research on Arabic diacritization was devoted to MSA for a number of reasons among which the availability of resources. Till recent, written dialects was very scarce. 
Since dialects have mostly eliminated case endings, we focus on word-internal diacritization.  Many approaches have been explored for word-internal diacritization of MSA such as Hidden Markov Models \cite{gal2002hmm,darwish2017arabic}, finite state transducers \cite{nelken2005arabic}, character-based maximum entropy based classification \cite{zitouni2006maximum}, and deep learning \cite{abandah2015automatic,belinkov2015arabic,rashwan2015}. \newcite{darwish2017arabic} compared their system to others on a common test set. They achieved a WER of 3.29\% compared 3.04\% for \newcite{rashwan2015}, 6.73\% for \newcite{habash2007arabic}, and 14.87 for \newcite{belinkov2015arabic}.  \newcite{azmi2015survey} survey much of the literature on MSA diacritization. For dialectal diacritization, the literature is rather scant.  \newcite{habash2012morphological} developed a morphological analyzer for dialectal Egyptian, which also performs diacritization using a finite state transducer that encodes manually crafted rules.  They report an overall analysis accuracy of 92.1\% without reporting diacrtization results specifically. \newcite{khalifa2017morphological} developed a morphological analyzer for dialectal Gulf verbs, which also attempts to recover diacritics. Again, they did not specifically report on diacritization results. \newcite{jarrar2017curras} annotated and diacritized a corpus of dialectal Palestinian containing 43k words.\citep{DARWISH18.20} used a collection of 8,200 verses from Moroccan and Tunisian dialectal Bible to build a Linear Chain CRF to recover word diacritics. They achieved a word level diacritization error of 2.9\%  and 3.8\% on Moroccan and Tunisian respectively.  

\section{Data}
We used the same data for ~\cite{DARWISH18.20} that is composed of two translations of the New Testament into two Maghrebi sub-dialects, namely Moroccan\footnote{Translated by Morocco Bible Society} and Tunisian\footnote{Translated by United Bible Societies, UK}. Both contains 8,200 verses each with 134,324 and 131,923 words for MOR and TUN respectively. 
Table \ref{table:bible-sample} gives a sample verse from both dialects with English translation. The data has two distinguishing properties, namely: it is religious in nature; and spelling is mostly consistent. Other dialectal text from social media differ in both of these aspects. For future work, we plan to extend this work to social media text.



\begin{table}
\begin{center}
\begin{tabular}{c|p{4cm}}
Lang. & Verse (Matthew 10:12) \\ \hline
MOR & \small \<وْإِلَا دْخَلْتُو لْشِي دَارْ، سَلّْمُو عْلَى مَّالِيهَا> \\
TUN & \small \<وكتُدْخْلُوا لْدَارْ سَلّْمُوا عْلَى النَّاسْ الِّي فِيهَا> \\
MSA & \small \< وَحِينَ تَدْخُلُونَ ٱلْبَيْتَ سَلِّمُوا عَلَيْهِ> \\
EN &  As you enter the home, greet those who live there \\
\end{tabular}
\caption{Sample verse from Bibles}
\label{table:bible-sample}
\end{center}
{\vspace{-8mm}}
\end{table}

We split the data 
into 5 folds for cross validation, where training splits were further split 70/10 for training/validation.  Given the training portions of each split, Table \ref{table:msa-variations} 
shows the distribution of the number of observed diacritized forms per word.  As shown, 89\% and 82\% of words have one diacritized form for MOR and TUN respectively. We further analyzed the words with more than one form.  The percentage of words where one form was used more than 99\% of time was 53.8\% and 55.5\% for MOR and TUN respectively. Similarly, the percentage of words where the most frequent form was used less than 70\% was 6.1\% and 8.5\% for MOR and TUN respectively.  We looked at alternative diacritized forms and found that the less common alternatives involve: omission of default diacritics (ex. \textit{fatha} before \textit{alef} -- {\small \<هاذا>} (hA*A) vs. {\small \<هَاذَا>} (haA*aA) -- ``this''); use of  \textit{shadda}--\textit{sukun} instead of \textit{sukun} (ex. {\small \<يِطَّهْرُوا>} (yiT$\sim$ahoruwA) vs. {\small \<يِطَّهّْرُوا>} (yiT$\sim$ah$\sim$oruwA) -- ``to purify''); use of alternative diacritized forms that have nearly identical pronunciation (ex. {\small \<نِكْذْبُوا>} (niko*obuwA) vs. {\small \<نْكَذّْبُوا>} (noka*$\sim$obuwA) -- ``we deny''); and far less commonly varying forms (ex. {\small \<قَلِّقْ>} (qal$\sim$iqo -- ``to cause anxiety'') -- vs. {\small \<قْلَقْ>} (qolaqo -- ``anxious'')).

\begin{table}[ht]
\begin{center}
\small
\begin{tabular}{c|c|c|c|c}
	& MOR & TUN & \multicolumn{2}{c}{MSA} \\
    & Bible	& Bible & Bible & News \\ \hline
Most Freq & 99.1 & 98.9 & 92.1 & 92.8 \\ \hline
\multicolumn{5}{c}{No. of Seen Forms} \\ \hline
1 & 89.0 & 81.8 & 51.7 & 69.0 \\ 
2 & 10.2 & 8.2 & 20.4 & 26.8 \\ 
3 & 0.8 & 5.6 & 13.5 & 2.9 \\ 
4 & 0.0 & 2.3 & 7.1 & 1.1 \\ 
$\geq$5 & 0.0 & 0.0 & 7.3 & 0.1 \\
\end{tabular}
\caption{Distribution of the number of dicaritized forms per word}
\label{table:msa-variations}
\end{center}
\end{table}

Further, we used the most frequent diacritized form for each word, and we automatically diacritized the training set (``Most Freq'' line in Table \ref{table:msa-variations}).  The accuracy was 99.1\% and 98.9\% for MOR and TUN respectively.  \textbf{This indicates that diacritizing words out of context may achieve up to 99\% accuracy.} We compared this to the MSA version of the same Bible verses (132,813 words) and a subset of diacrtized MSA news articles of comparable size (143,842 words) after removing case-endings.  As Table \ref{table:msa-variations} shows, MSA words, particularly for the Bible, have many more possible diacritized forms, and picking the most frequent diacritized form leads to significantly lower accuracy compared to dialects. 

We compared the overlap between training and test splits.  
We found that 93.8\% and 93.4 of the test words were observed during training for MOR and TUN respectively.  If we use the most frequent diacritized forms observed in training, we can diacritize 92.8\% and 92.0\% of MOR and TUN words respectively.  Thus, the job of a diacritizer is primarily to diacritize words previously unseen words, rather than to disambiguate between different forms.  We also compared the cross coverage between the MOR and TUN datasets.  
The overlap is approximately 61\%, and the diacritized form in one dialect matches that of another dialect less than two thirds of the time.  This suggests that cross dialect training will yield suboptimal results. Other notable aspects of MOR and TUN that set them apart from MSA are: both allow leading letters in words to have \textit{sukun} (MOR: 34\% and TUN: 26\% of words); MOR uses a \textit{shadda}--\textit{sukun} combination; and both allow consecutive letters to have \textit{sukun} (ex. all letter in the MOR word {\small \<وْلْلْبْلَايْصْ>} (wololobolaAyoSo -- ``and places'') have \textit{sukun} save one).

\section{Proposed Approach}

\begin{figure}[h]  
\begin{center}
\includegraphics[scale=0.08]{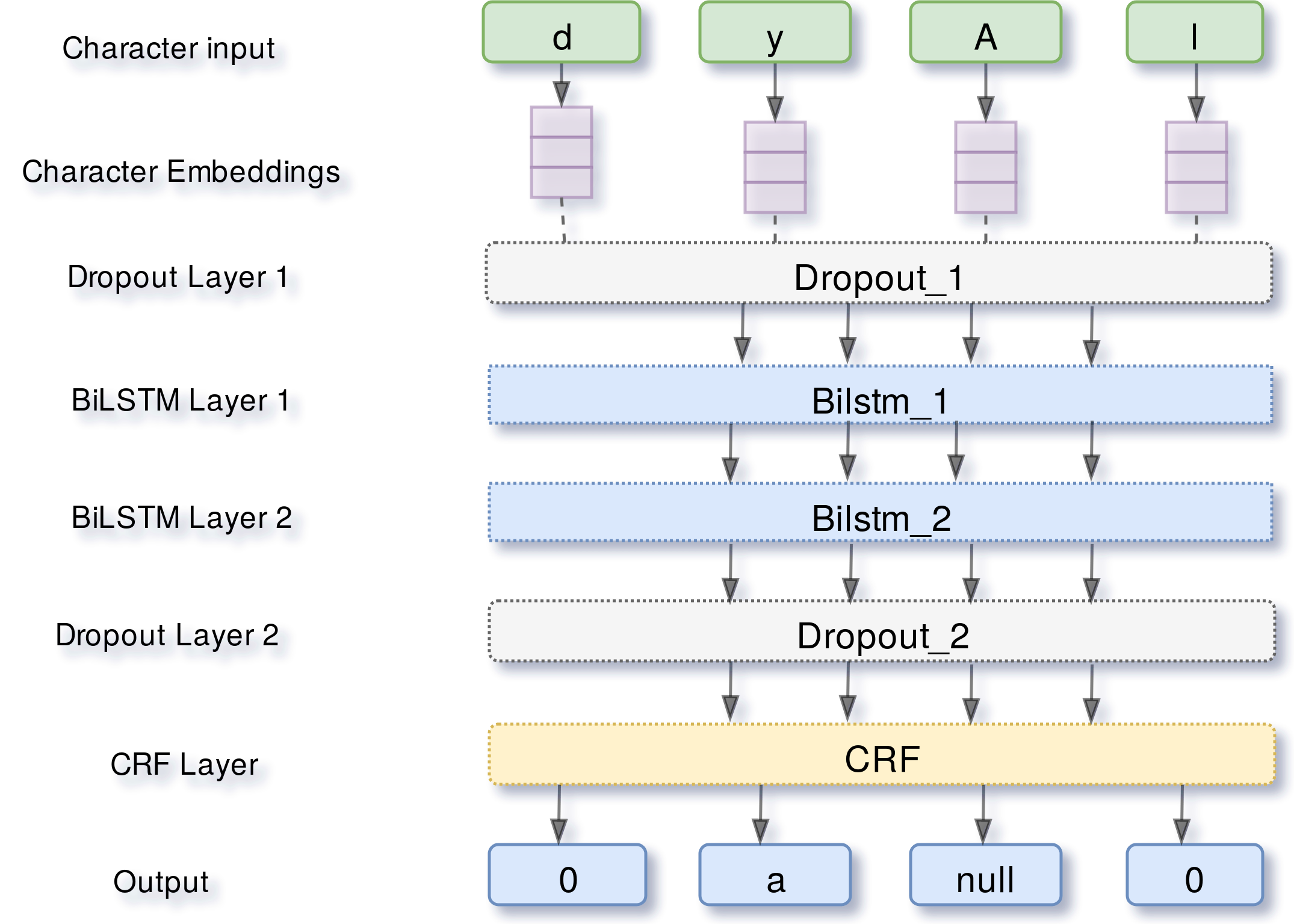}
\caption{DNN architecture}
\label{figure:DNNdiacarch}
\end{center}
\end{figure}

Capitalizing on the success of neural approaches ~\cite{belinkov2015arabic,abandah2015automatic} and more precisely biLSTMs and CRF~\cite{lample16,ling2015not}, 
 we implemented the architecture shown in Figure \ref{figure:DNNdiacarch} with four layers: one input, one output, and two hidden layers. At the input, a look-up table of randomly initialized embeddings maps each input character to a d-dimensional vector. The output from the character fixed-dimensional embeddings is used as input to the two hidden layers containing two stacked Bidirectional Long Short Term Memory (biLSTM)~\cite{schuster1997bilstm} layers.  BiLSTMs have shown their effectiveness in processing sequential data as they capture  long-short term dependencies within the characters~\cite{graves2012supervised}. At the output layer, a CRF layer is applied over the hidden representation of the two stacked biLSTMs to obtain the probability distribution over all labels. Since biLSTMs produce probability distribution for each output independently from other outputs, CRFs help overcome this independence assumptions and impose sequence labeling constraints. In our scenario, this was 12 possible tags representing one of the possible diacritics or none. 
We used Adam~\cite{KingmaB14Adam} to optimize for the cross entropy objective function. Side experiments with stochastic gradient descent with momentum, AdaDelta ~\cite{zeiler2012adadelta}, and RMSProp~\cite{dauphin2015equilibrated} did not lead to improvements. 
To avoid overfitting, we applied dropout~\cite{srivastava2014dropout} and early stopping.  
Dropout prevents co-adaptation of the hidden units by randomly setting a portion of hidden units to zero during training. 
We used early stopping with patience equal to 10. 
If validation error did not improve enough after this number of times, training is stopped. 
We tuned hyper-parameters on the development dataset by using random search resulting in the following parameters: 

\begin{table}[h]
\begin{center}
\begin{tabular}{llp{1cm}p{1cm}p{1cm}p{1cm}p{1cm}p{1cm}p{1cm}|}
\hline
Layer &  \multicolumn{1}{l}{Hyper-Parameters} & Value  \\ \hline

\multirow{2}{*}{Bi-LSTM} 
                      & state size     & 200 \\   
                      & initial state  & 0.0  \\ \hline
              Dropout & dropout rate    & 0.25 \\ \hline
              Characters Emb. & dimension  & 100 \\ \hline
\multirow{3}{*}{} 
                      & batch size     & 5 \\   
                      & learning rate  & 0.01  \\
                      & decay rate  & 0.05  \\ \hline
\end{tabular}
\label{table:hyper}
\end{center}
\end{table}

The baseline results provided by ~\citep{DARWISH18.20} used CRF sequence labeling~\cite{lafferty2001conditional}, which has shown effectiveness for many sequence labeling tasks. CRFs effectively combine state-level and transition features. 
CRF++ implementation of a CRF sequence labeler with L2 regularization and default value of 10 for the generalization parameter ``C''\footnote{\url{https://github.com/taku910/crfpp}}  with letters as the inputs and per letter diacritics as labels was used. For features, given a word of character sequence $c_n$ ... $c_{-2}$, $c_{-1}$, $c_0$, $c_1$, $c_2$ ... $c_m$, we used a combination of character n-gram features, namely unigram ($c_0$), bigrams ($c_{-1}^0$; $c_0^{1}$), trigrams ($c_{-2}^0$; $c_{-1}^{1}$; $c_0^{2}$), 4-grams ($c_{-3}^0$; $c_{-2}^{1}$; $c_{-1}^{2}$; $c_{0}^{3}$), and Brown Clusters~\cite{brown1992class}.  Given that the vast majority of dialectal words have only one possible diacritized form, the CRF is trained on individual words out of context.  ~\cite{DARWISH18.20}.

\section{Results and Discussion}
We conducted three sets of experiments in contrast to previous results using CRF
which
achieved WER of 3.1\% and 4.0\% for MOR and TUN respectively (Table \ref{table:CRFresultsBrownClusters}). 
The DNN model edged the CRF approach with 0.4\% drop in WER for both dialects (Table \ref{table:DNNresults} (a)). 

\begin{table}[ht]
\begin{center}
\begin{tabular}{c|c|c|c}
		&			& \multicolumn{2}{c}{Error Rate} \\ 
Training Set	&	Test Set	&	Character	&	Word	\\ \hline
\multicolumn{4}{c}{(a) Uni-dialectal Training} \\ \hline
Moroccan	&	Moroccan	&	1.1	&	\textbf{2.9}	\\
Tunisian	&	Tunisian	&	1.7	&	\textbf{3.8} \\ \hline
\multicolumn{4}{c}{(b) Cross Training} \\ \hline
Moroccan	&	Tunisian	&	20.1	&	47.0	\\
Tunisian	&	Moroccan	&	20.8	&	48.9	\\ \hline
\multicolumn{4}{c}{(c) Combined Training} \\ \hline
Combined	&	Moroccan	&	12.6	&	\textbf{34.2}	\\ 
Combined	&	Tunisian	&	9.5	&	\textbf{23.8}	\\
\end{tabular}
\caption{CRF Results with Brown clusters reported by ~\cite{DARWISH18.20}}
\label{table:CRFresultsBrownClusters}
\end{center}
\end{table}

\begin{table}[ht]
\begin{center}
\small
\begin{tabular}{c|c|c|c}
		&			& \multicolumn{2}{c}{Accuracy} \\ 
Training Set	&	Test Set	&	Character	&	Word	\\ \hline
\multicolumn{4}{c}{(a) Uni-dialectal Training} \\ \hline
MOR	&	MOR	&	\textbf{1.0}	&	\textbf{2.7}	\\
TUN	&	TUN	&	\textbf{1.6}	&	\textbf{3.6}	\\ \hline
\multicolumn{4}{c}{(b) Cross Training} \\ \hline
MOR	&	TUN	&	21.4	&	48.2	\\
TUN	&	MOR	&	22.3	&	49.4	\\ \hline
\multicolumn{4}{c}{(c) Combined Training} \\ \hline
Joint	&	MOR	&	\bf{1.3}	&	\textbf{3.7}	\\ 
Joint	&	TUN	&	\bf{2.1}	&	\bf{4.9}	\\  \hline
\end{tabular}
\caption{DNN Results -- Average Across All Folds}
\label{table:DNNresults}
\end{center}
\end{table}
Second, we tested if sub-dialects can learn from each other.  Tables \ref{table:CRFresultsBrownClusters} (b) and \ref{table:DNNresults} (b) show that cross-dialectal results were significantly lower than mono-dialectal ones, 
confirming that dialects are phonetically divergent. 
Identical words with different diacritized forms in both dialects abound. Examples include {\small \{
\<مَنْطْقَةْ> (manoToqapo), 
\<مْنْطِقَةْ> (manoTiqapo)\}} (region)  and {\small \{
\<طْفُلْ> (Tofulo), 
\<طْفَلْ> (Tofalo)\}} (boy) in MOR and TUN respectively. 

Third, we combined training data from both dialects, and we tested on individual dialects. Tables \ref{table:CRFresultsBrownClusters} (c) and \ref{table:DNNresults} (c) show the results of joint training. While the CRF baseline results were significantly worse, DNN WER increased by 1\% and 1.3\% for MOR and TUN respectively. The results suggest that unlike CRFs, our DNN model was implicitly identifying the sub-dialect. 
\begin{table}[ht]
\begin{center}
\small
\begin{tabular}{c|c|c|l}
\hline
T & P & R & Examples \\
\hline
\multicolumn{4}{c}{MOR}  \\
\hline
$\sim$  & $\sim$u & 8.8\% & \<التّبَهْ>$\rightarrow$\<التُّبَهْ> ``the hill'' \\
	& & & (Alt\textbf{$\sim$}baho$\rightarrow$Alt\textbf{$\sim$u}baho) \\
$\sim$a & $\sim$u & 2.5\% & \<الصَّدَقَة>$\rightarrow$\<الصُّدْقَة> ``the charity'' \\
	& & & (AlS\textbf{$\sim$a}daqap$\rightarrow$AlS\textbf{$\sim$u}daqap) \\ 
o & $\sim$ & 3.4\% &  \<تْعَطّلْ>$\rightarrow$\<تّعَطّلْ> ``delays'' \\
	& & & (t\textbf{o}EaT$\sim$lo$\rightarrow$t\textbf{$\sim$}EaT$\sim$lo) \\
\hline
\multicolumn{4}{c}{TUN} \\
\hline
o & $\sim$ & 14.3\% &  \<الجْمَالْ>$\rightarrow$\<الجّمَالْ> ``camels'' \\
	& & & (Alj\textbf{o}mal$\rightarrow$Alj\textbf{$\sim$}mal) \\
$\sim$i & $\sim$ & 2.1\% &  \<غَلِّتْهَا>$\rightarrow$\<غَلّتْهَا> ``its fruits'' \\
	& & & (gal\textbf{$\sim$i}toha$\rightarrow$gal\textbf{$\sim$}toha) \\
a & $\sim$a & 3.0\% &   \<ضْيَافْ>$\rightarrow$\<ضَيَّافْ> ``guests'' \\
	& & & (D\textbf{o}y\textbf{a}Af$\rightarrow$D\textbf{a}y\textbf{$\sim$a}Af) \\
\hline
\end{tabular}
\caption{Most common diacritic prediction errors (T: Truth, P: Predicted, R: Ratio)} 
\label{table:dia-common-errors}
\end{center}
\end{table}

\begin{figure}[h]  
\begin{center}
\includegraphics[scale=0.42]{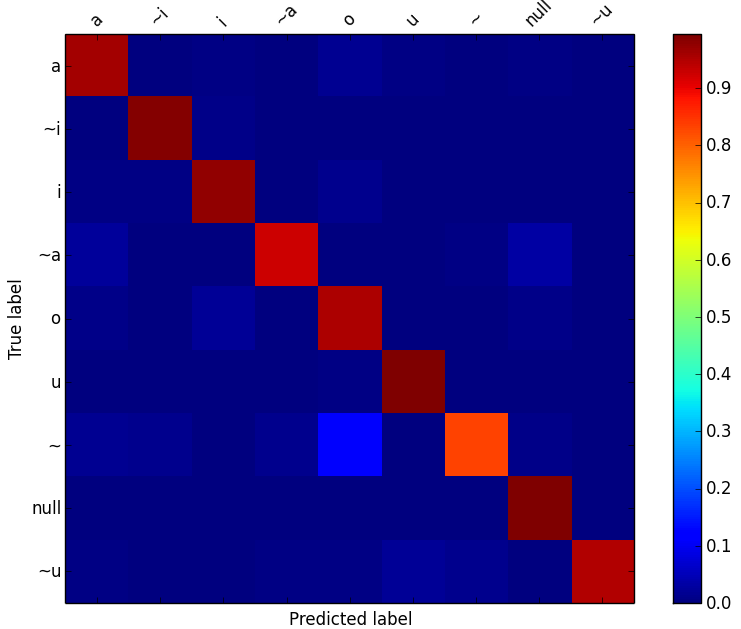}
\caption{Confusion Matrix for Moroccan using Combined approach.}
\label{figure:morcomb}
\end{center}
\end{figure}

\begin{figure}[h]  
\begin{center}
\includegraphics[scale=0.42]{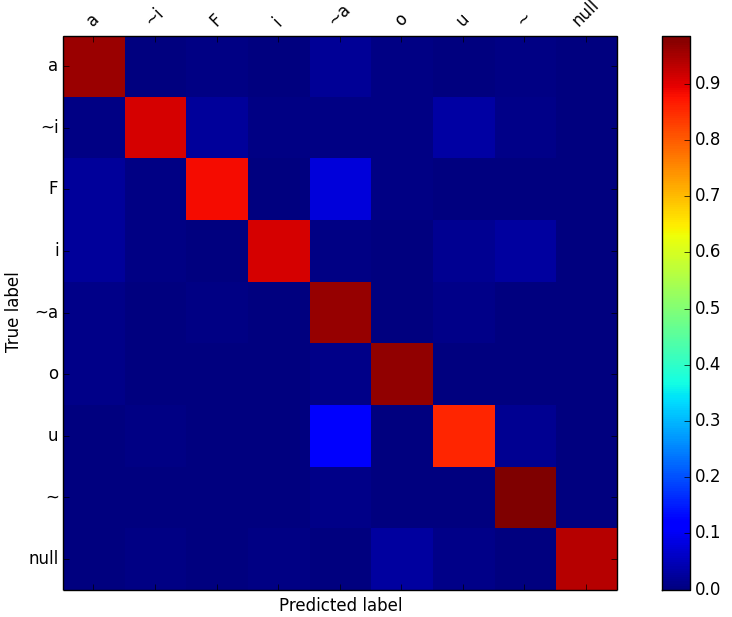}
\caption{Confusion Matrix for Tunisian using Combined approach.}
\label{figure:tuncomb}
\end{center}
\end{figure}

Figures \ref{figure:morcomb} and \ref{figure:tuncomb} displays the combined model confusion matrices. While both figures shows that the joint model was able to predict accurately the correct diacritic (label); Some errors can be noted, manly errors involve \textit{shadda} ($\sim$) or   \textit{sukun} (o); 
Table \ref{table:dia-common-errors} details the most common errors for both sub-dialects with error examples.  The most common errors involved \textit{fatha} (a), \textit{shadda} ($\sim$), \textit{sukun} (o), and kasra (i). We also looked at the percentage of errors for individual diacritics (or combinations in which they appear) using mono-dialectal and joint training.  The break-down was as follows:
\begin{table}[ht]
\begin{center}
\begin{tabular}{c|c|c|c|c}
	&	\multicolumn{2}{c}{MOR}			&	\multicolumn{2}{|c}{TUN}			\\
	&	Mono	&	Joint	&	Mono	&	Joint	\\ \hline
fatha (a)	&	68.4	&	58.1	&	83.0	&	51.0	\\
sukun (o)	&	63.3	&	64.9	&	55.8	&	56.1	\\
shadda ($\sim$)	&	48.5	&	42.7	&	30.4	&	29.0	\\
kasra (i)	&	14.6	&	14.6	&	52.7	&	43.7	\\
damma (u)	&	11.8	&	13.4	&	20.5	&	19.1	\\\hline
\end{tabular}
\end{center}
\end{table}

The breakdown shows that error types in MOR and TUN were rather different.  For example, \textit{kasra} error were more pronounced in TUN than MOR. Also, joint training affected different diacritics differently.  For example, joint training for TUN caused a very large drop in errors for \textit{fatha}.

Given our results, we suggest that an effective strategy for robust dialectal diacritization would involve: a) building, with the help of our model, a large lookup table for the most common words with one possible diacritized form for each dialect, which would account for 99\% of the words, and using simple lookup for seen words in the lookup table and using the diacritization model otherwise; and b) using a mono-dialectal model in application where the sub-dialect is known (ex. chat app in a specific country) and resorting to the combined model otherwise (ex. tweets of unknown source).

\section{Conclusion}
In this paper, we have presented the diacritization of Maghrebi Arabic, a dialect family used in Northern Africa. This work will help enable NLP to model conversational Arabic in dialog systems. We noted that dialectal Arabic is less contextual and more predictable than Modern Standard Arabic, and high levels of accuracy can be achieved if enough data is available. We used a character-level DNN architecture that stacks two biLSTM layers over a CRF output layer. Mono-dialectal training achieved WER less than 3.6\%. Though sub-dialects are phonetically divergent, our joint training model implicitly identifies sub-dialects, leading to small increases in WER.
\bibliography{diacritization-maghrebi-arabic}
\bibliographystyle{acl_natbib}

\end{document}